\documentclass[conference]{IEEEtran}
\IEEEoverridecommandlockouts

\usepackage{cite}
\usepackage{times}  
\usepackage{helvet}  
\usepackage{courier}  
\usepackage[hyphens]{url}  
\usepackage{graphicx} 

\usepackage{caption} 
\usepackage{amsmath,amssymb,amsfonts}
\usepackage{graphicx}
\usepackage{textcomp}
\usepackage{xcolor}
\usepackage{amsthm,amsmath,amssymb}
\usepackage{mathrsfs}
\usepackage{booktabs}
\usepackage{verbatim}
\usepackage{siunitx}
\usepackage{amssymb}
\usepackage{threeparttable}

\usepackage{algorithm}
\usepackage{algorithmic}
\usepackage{amssymb}
\usepackage{pifont}
\usepackage{subcaption}

\usepackage{newfloat}
\usepackage{listings}
\def\BibTeX{{\rm B\kern-.05em{\sc i\kern-.025em b}\kern-.08em
    T\kern-.1667em\lower.7ex\hbox{E}\kern-.125emX}}

\title{A Versatile Agent for Fast Learning from Human Instructors}

\author {
    Yiwen Chen$^{\dag *}$,
    Zedong Zhang$^{\dag}$,
    Haofeng Liu, 
    Jiayi Tan,
    Chee-Meng Chew,
    Marcelo Ang
\\Department of Mechanical Engineering, National University of Singapore

\thanks{$\dag$ These authors contributed equally.}
\thanks{$^{*}$Corresponding author: Yiwen Chen  (yiwen.chen@u.nus.edu).}

\thanks{This research is supported by the Agency for Science, Technology and Research (A$^{*}$STAR), Singapore, under its AME Programmatic Funding Scheme (Project $\#$A18A2b0046). The computational work for this article was partially performed on resources of the National Supercomputing Centre, Singapore (https://www.nscc.sg).}
}

\begin{document}
\maketitle

\begin{abstract}
In recent years, a myriad of superlative works on intelligent robotics policies have been done, thanks to advances in machine learning. However, inefficiency and lack of transfer ability hindered algorithms from pragmatic applications, especially in human-robot collaboration, when few-shot fast learning and high flexibility become a wherewithal. To surmount this obstacle, we refer to a "Policy Pool", containing pre-trained skills that can be easily accessed and reused. An agent is employed to govern the "Policy Pool" by unfolding requisite skills in a flexible sequence, contingent on task specific predilection. This predilection can be automatically interpreted from one or few human expert demonstrations. Under this hierarchical setting, our algorithm is able to pick up a sparse-reward, multi-stage knack with only one demonstration in a Mini-Grid environment, showing the potential for instantly mastering complex robotics skills from human instructors. Additionally, the innate quality of our algorithm also allows for lifelong learning, making it a versatile agent.
\end{abstract}

\section{Introduction}
The prosperity of deep learning in the last decade has led to numerous conspicuous achievements in intelligent agents. Especially in the realm of reinforcement learning (RL), works such as mastering the game of Go \cite{silver2016mastering}, playing video games \cite{mnih2015human}, robotics control \cite{amarjyoti2017deep}, etc. are noteworthy. 

Although the synergy of deep neural networks and novel learning algorithms makes powerful agents, the lack of learning efficiency is an unsolved problem. It was shown that in playing Atari games, intelligent artificial agents require more learning samples than humans by several orders of magnitude \cite{tsividis2017human}. This phenomenon is essentially a corollary of ineffectiveness in transfer ability, which means that the agent has to learn each new task from scratch.

The aforementioned problem can largely limit the competence of deep learning algorithms when learning efficiency is a must. For example, a recent hot topic, Interactive Task Learning (ITL), broaches a plausible future scenario in which intelligent artificial agents learn from human instructors \cite{laird2017interactive}. In this case, it is impractical for human instructors to teach hundreds of times, and thus one-shot or few-shot learning is expected.

\begin{figure}[h]
    \centering
     \includegraphics[width=0.47\textwidth]{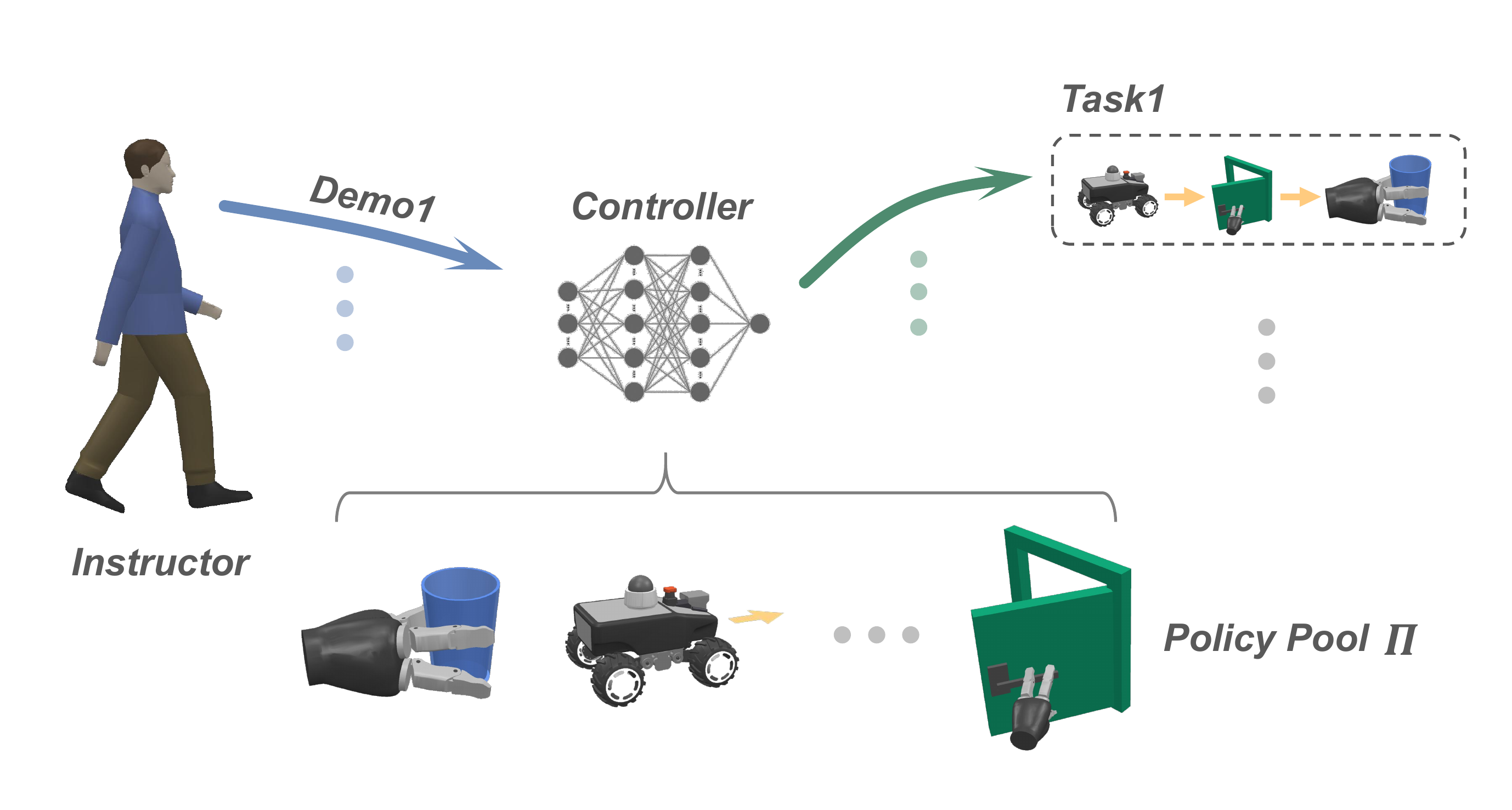}
    \caption{A schematic representation of our proposed method.}
    \label{fig:intro}
\end{figure}

In this work, we focus on the following task: Intelligent artificial agents' fast learning from demonstrations of human instructors. We argue that the quintessence of tackling this task is threefold: (1)  \textbf{Efficiency:} The agent should have a few-shot learning ability. (2) \textbf{Inclusiveness:} The agent can handle a wide range of tasks, including sparse-reward, multi-stage tasks. (3)  \textbf{Flexibility:} The agent should be flexible and can adapt to tweaks of the learned policies. The significance of the first point has already been delineated previously. Regarding inclusiveness, we reckon that the agent should be able to cover disparate types of task, so that it can respond to broader instructions. Sparse-reward, multi-stage, or long sequence tasks have long been a conundrum for RL due to the imbalance in reward distribution. If the agent can overcome this problem, it can be more inclusive, particularly for complex tasks. Moreover, we believe that the essence of learning from humans is high task flexibility, otherwise a pre-programmed agent would be preferred. In this sense, the agent should be robust and adapt to new task preferences, even if it is different from the tasks on which it was initially trained.

To fulfill the three requirements, one solution is through policy reuse, i.e., employing properly defined sub-skills that exhibit recurrent pattern in solving complex tasks. This idea is similar to the theory of hierarchical reinforcement learning (HRL). Previous works claim that this is a way of decomposing complex tasks and solving them by divide-and-conquer \cite{barreto2020fast,tessler2017deep}. Meanwhile, from the perspective of lifelong learning, it is likely that in the future a skill library could be available in which deep learning models are used to represent fine-tuned canonical skills \cite{chen2018lifelong}. Then the idea becomes solving non-trivial tasks by the synergy of trivial skills. All in all, either opinion endorses the idea of policy reuse. 

As illustrated in Figure \ref{fig:intro}, our proposed method introduces a Policy Pool $\Pi$, which contains pre-trained sub-skills. An agent is then employed as a controller of the Policy Pool. The controller takes in human demonstrations and learns a task predilection, which guides the agent to form a higher policy for the new task. For instance, the agent can generate a policy of navigating to a door, opens that door, and grasps the cup behind the door if the corresponding skills are pre-trained and accessible to the controller. This architecture is analogous to some HRL algorithms except that we can achieve few-shot learning, which is realized through observation minimization and reward-augmented imitation learning (see Approach Section). 

Briefly, we introduce our work as Fast Imitation and Policy Reuse Learning (FIRL).
The FIRL agent can learn much faster than conventional HRL, while, when compared with linear policy combination methods, our agent has the advantage of automatically interpreting task predilections from demonstrations. We elucidate in the experiment that our agent can learn a three-stage task with only one demonstration in a Mini-Grid environment, showing the potential of mastering complex robotics policies instantly. Lastly, the innate quality of the FIRL algorithm can naturally support lifelong learning. A new sub-skill can be added to the Policy Pool at anytime, which takes effect in no time.

\section{Related Works}
Essentially, there are three fields of study that are related to our work, viz. hierarchical reinforcement learning, lifelong learning, and rapid learning in few shot, among which there are two works that share a most similar spirit with us. The first work is proposed by Barreto et al. \cite{barreto2019option,barreto2020fast}, which is based on Successor Features \cite{barreto2017successor}. Their salient point is to assign weights to value functions of pre-trained sub-skills, then new policies can be formed according to a generalized policy update trick. This algorithm is especially suitable for cases where the reward function for the new task is a linear combination of sub-skills' reward functions. In that case, the aforementioned weights are the coefficients of the linear combination, which can be acquired directly to realize fast and flexible learning. Albeit a seminal work, its mechanism requires all sub-skills to have the same action space, and lacks the ability in solving multi-stage tasks. The other work is called H-DRLN \cite{tessler2017deep}. Although a similar architecture can be observed, the work fixates on lifelong learning, while its learning efficiency is not accentuated. Thus, these previous works are not very useful for learning from human demonstrations.

The background of literature of the three indicated fields will also be introduced: \textbf{Hierarchical reinforcement learning:}
HRL endorses the concept of temporal abstraction and options. Fundamental HRL algorithms include HAMs \cite{parr1997reinforcement}, MAXQ \cite{dietterich1998maxq,dietterich2000hierarchical} and option framework \cite{sutton1999between}. Based on these precedents, various HRL methods \cite{kulkarni2016hierarchical,nachum2018data} have been proposed to solve long-horizon tasks by reducing search space. Options discovering methods were introduced \cite{krishnan2017ddco,fox2017multi}, specifically, Kipf et al. \cite{kipf2019compile} proposed
an encoder-decoder model for skills extraction and composition. Furthermore, \cite{gupta2019relay} shows the capability to extract skills abstraction from demonstration dataset. Skills priors based methods \cite{hakhamaneshi2021hierarchical,hausman2018learning,pertsch2020accelerating}, analogous to options discovering, extract skill representation from demonstrations dataset. 

\textbf{Lifelong learning:} Lifelong learning (a.k.a. Continual learning) is an integral topic, whose incentive is to save the cost of repetitive training and to ensure that skills could continue to expand without catastrophic forgetting \cite{chen2018lifelong,li2019learn}. The branches of study include regularization based methods \cite{kirkpatrick2017overcoming,zenke2017continual}, parameter isolation approaches \cite{rusu2016progressive} and replay based methods \cite{lopez2017gradient,rebuffi2017icarl}.

\textbf{Few-shot learning:} Fast learning methods allow agent to rapidly master new skills. One-shot and few-shot learning are similar topics related
to fast learning that requires minimal training samples \cite{hakhamaneshi2021hierarchical}. Duan et al. gave convincing results for one-shot imitation learning
using context and attention extraction \cite{duan2017one}.

\section{Approach}
\begin{figure*}[h]
    \centering
     \includegraphics[width=0.98\textwidth]{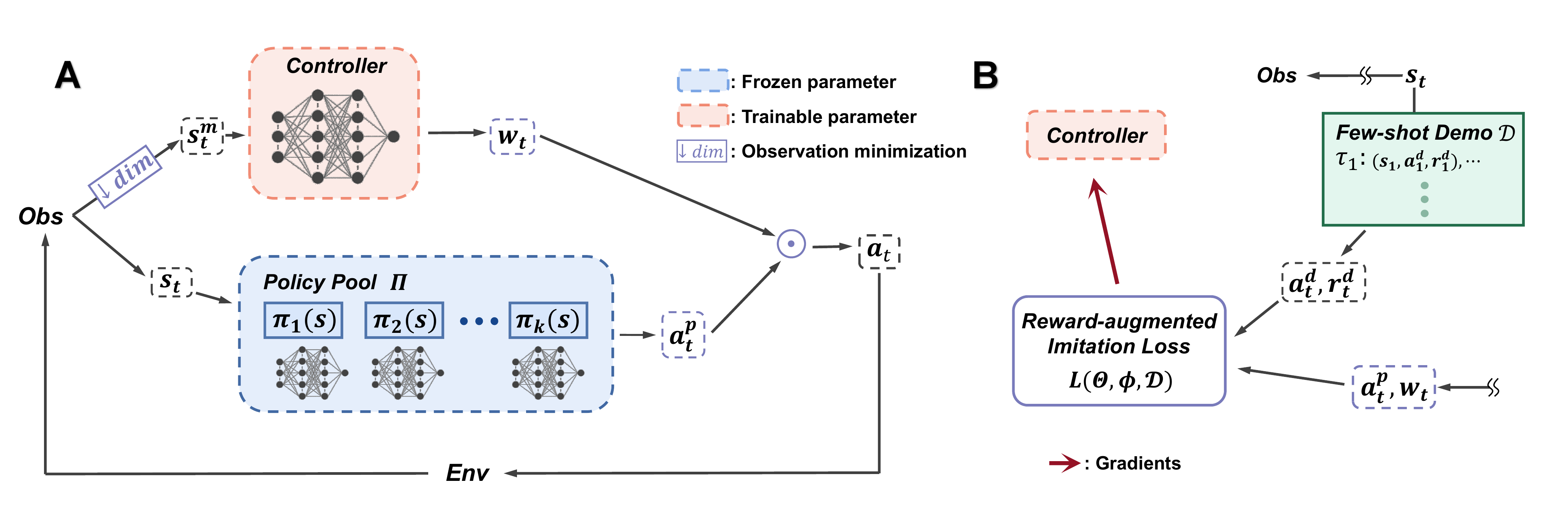}
    \caption{Details of the FIRL architecture. A: During evaluation, the agent interacts with the environment and receives an observation. The vanilla observation will be fed to the sub-policies in the Policy Pool, while a reduce-dimension observation, i.e., $s^{m}_{t}$ is forwarded to the controller. The controller outputs a task predilection weights $w_t$ for choosing sub-skills. B: During training, the agent learns from demonstrations. At each time step, and actions from sub-policies $a^{p}_{t}$ are compared with action from demonstration $a^{d}_{t}$. The result along with the predilection vector, reward $r^{d}_{t}$ are used in computing the imitation loss. Lastly, the parameters of the controller is updated.}
    \label{fig:architecture}
\end{figure*}

The objective of FIRL is to enable fast learning from human demonstrations while accessing a Policy Pool $\Pi$. How we achieve efficient and flexible learning is through policy reuse, i.e., building a hierarchical policy architecture. Furthermore, how we promote the efficient learning to few-shot learning is through a reward-augmented imitation learning mechanism, along with observation minimization. The main architecture of our approach is shown in Figure \ref{fig:architecture}.

\paragraph{Problem Formulation}
Assume that we have access to a Policy Pool $\Pi$ as well as a few-shot demonstration \(\mathcal{D}\). Policy pool  \(\Pi=\{\pi_{\theta_1}(a|s),...,\pi_{\theta_k}(a|s)\}\) is a set of pre-trained skills. \(k\) is the number of skills. The demonstration set \(\mathcal{D}=\{\tau_{i},...,\tau_{N}\}\) contains a number of trajectories denoted by $\tau$. Compared to benchmark imitation learning methods \cite{pertsch2020accelerating,duan2017one}, we only have a limited number of trajectories. A trajectory encompasses a time series of tuples \(\tau_{i}=\{(s_{1},a_{1}^{d},r_{1}^{d}),...,(s_{T_{i}},a_{T_{i}}^{d},r_{T_{i}}^{d})\}\), each containing the observation (or in a fully observed case, the state), the action, and a reward. The notation \(d\) implies that the data is from demonstration and \(T_i\) is the episode length of trajectory \(i\). This dataset is collected from limited interactions with human instructors. \(N\) is the number of demonstration trajectories. 

\paragraph{Training of the sub-skills}
The sub-skills in the Policy Pool are pre-trained using the RL algorithm. Specifically, the process is modeled by the Markov Decision Process (MDP). An MDP denotes a tuple $M\equiv (\mathcal{S},\mathcal{A},p,r,\gamma)$ where $\mathcal{S}$ and $\mathcal{A}$ represent the state space and the action space, respectively. $p(\cdot |s,a)$ denotes the probability of transition of taking action $a\in \mathcal{A}$ in state $s\in \mathcal{S}$. The reward function is given by $r(s,a,s^{'})$, while $\gamma \in [0,1)$ is the discount factor. We aim to train a policy \(\pi_{\theta}(a|s)\) to maximize the expectation of decayed rewards \(E_{\pi}\left[\sum^{T-1}_{t=0}\gamma^{t}r_t(s_t,a_t)\right]\), where $\theta$ denotes the parameters for sub-skills. The algorithm we use is Proximal Policy Optimization (PPO) \cite{schulman2017proximal}.

\paragraph{Reward-augmented Imitation Learning} To achieve high training efficiency, we refer to imitation learning in controller training. In this phase, the parameters (weights) of the sub-skills are frozen. At each time step, the Policy Pool provides a series of primitive actions (as shown in Figure \ref{fig:architecture}) $a^{p}_{t} = [a^{p}_{t,1},..., a^{p}_{t,k}]$. The controller's mission is to produce a weight vector $w_t$ to select appropriate primitive actions.
\begin{equation}
    w_{t}=(w_{t,1},...,w_{t,k})^T\sim\pi_\phi(\cdot|s_{t}^{m})
\label{eq:1}
\end{equation}
where $\phi$ denotes the controller parameters. Although there could be better ways to make a collaborative decision, in this work, we employ a simple mechanism by taking the highest weight primitive action as the synthesized action of the agent.
\begin{equation}
    a_t\sim\pi_{\theta_{i}}(\cdot|s_t),\ i=argmax(w_t)
\end{equation}

Based on the previous discussion, imitation loss is defined as follows.
\begin{equation}
    L(\Theta,\phi,\mathcal{D})=
\frac{1}{T}\sum^T_{t=1} \big( \sum^k_{i=1}(w_{t,i}-\mathcal{I}(a_{t,i},a_{t}^{d}))^2 \big)
\end{equation}
where $\mathcal{I}$ is a calculation we have defined. It equals 1 when the discrepancy between two elements is within a small threshold. Otherwise, it equals 0.

With this imitation learning algorithm, efficient learning is possible. However, it still failed to live up to the objective of few-shot training. Thus, we introduce a modified version, that is, reward-augmented imitation loss $L^{'}$. Our motivation is to avoid the noise and over-fitting introduced by imitation learning by modulating the imitation loss. Specifically, we assign a higher weight to the loss function if the current step is significant in the trajectory, that is, it can elicit a high reward. 
\begin{equation}
    L^{'}=
\frac{1}{T}\sum^T_{t=1} r_{t}^{d} \big( \sum^k_{i=1}(w_{t,i}-\mathcal{I}(a_{t,i},a_{t}^{d}))^2 \big)
\end{equation}
Note that this reward $r^{d}$ is different from all sub-skills. It can be easily defined by users, so long as it follows the theory of RL. It can even be highly sparse as the training only partially relies on this reward. It is not typical for supervised learning methods to have a reward function. However, we found that this mechanism is a key in promoting learning efficiency (see Experimental Section, Ablation Study).

\begin{table}[h]
\caption{Example of observation minimization.}
\label{table:obs_example}
\begin{center}
\begin{tabular}{c||c|c}
\hline
Observation &\(s^{m}_{t}\) & \(s_t\) \\
\hline
Sparsity & sparse & dense \\
\hline
Dimension & 3 & 147 \\
\hline
Explanation & flags of door's open/close
& raw image \\
\hline
Example &
$[0,1,0]^T$
&
\begin{minipage}{.1\textwidth}
  \includegraphics[width=\linewidth]{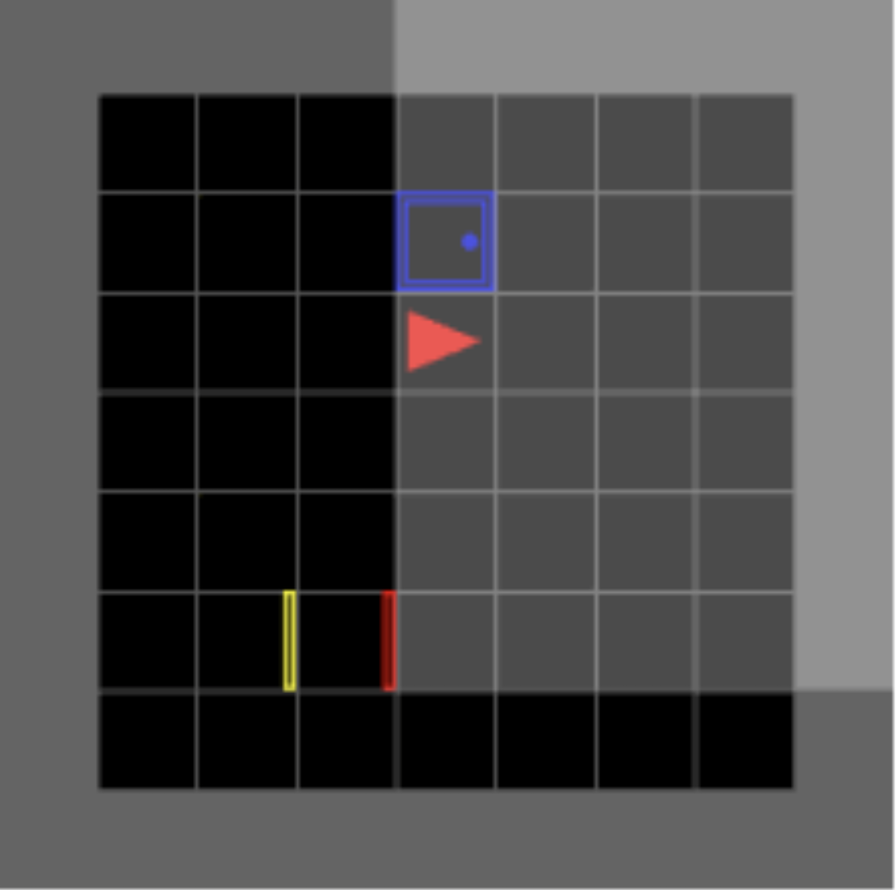}
\end{minipage}\\
\hline
\end{tabular}
\end{center}
\end{table}

\paragraph{Observation Minimization}
A straightforward approach to promote learning efficiency would be the reduction of observation. With our hierarchical architecture, a minimal observation dimension can be achieved without impairing the performance. Thus, we only retain the necessary features $s^{m}_{t}$ as an observation for the controller, as shown in Eq. \ref{eq:1}.

In Table \ref{table:obs_example}, we give a concrete example of what the necessary features may look like. In the example, it can be observed that the reduction is realized from the 147-dimensional original observation, i.e., the rgb image, to 3 using only the flags of interest, and the reduction can be more significant if we encounter a larger image.

\section{Experiment}
In this section, we answer the following questions:
\begin{itemize}
\item Could FIRL succeed in learning from a few-shot demonstrations? 
\item How is the performance improved compared to imitation learning method BC (Behavior Clone) and nonimitation learning methods, e.g., PPO and APEX-DQN?
\item Can we further improve performance?
\end{itemize}
\paragraph{Environment Setup}
\begin{figure}[h]
    \centering
     \includegraphics[width=0.3\textwidth]{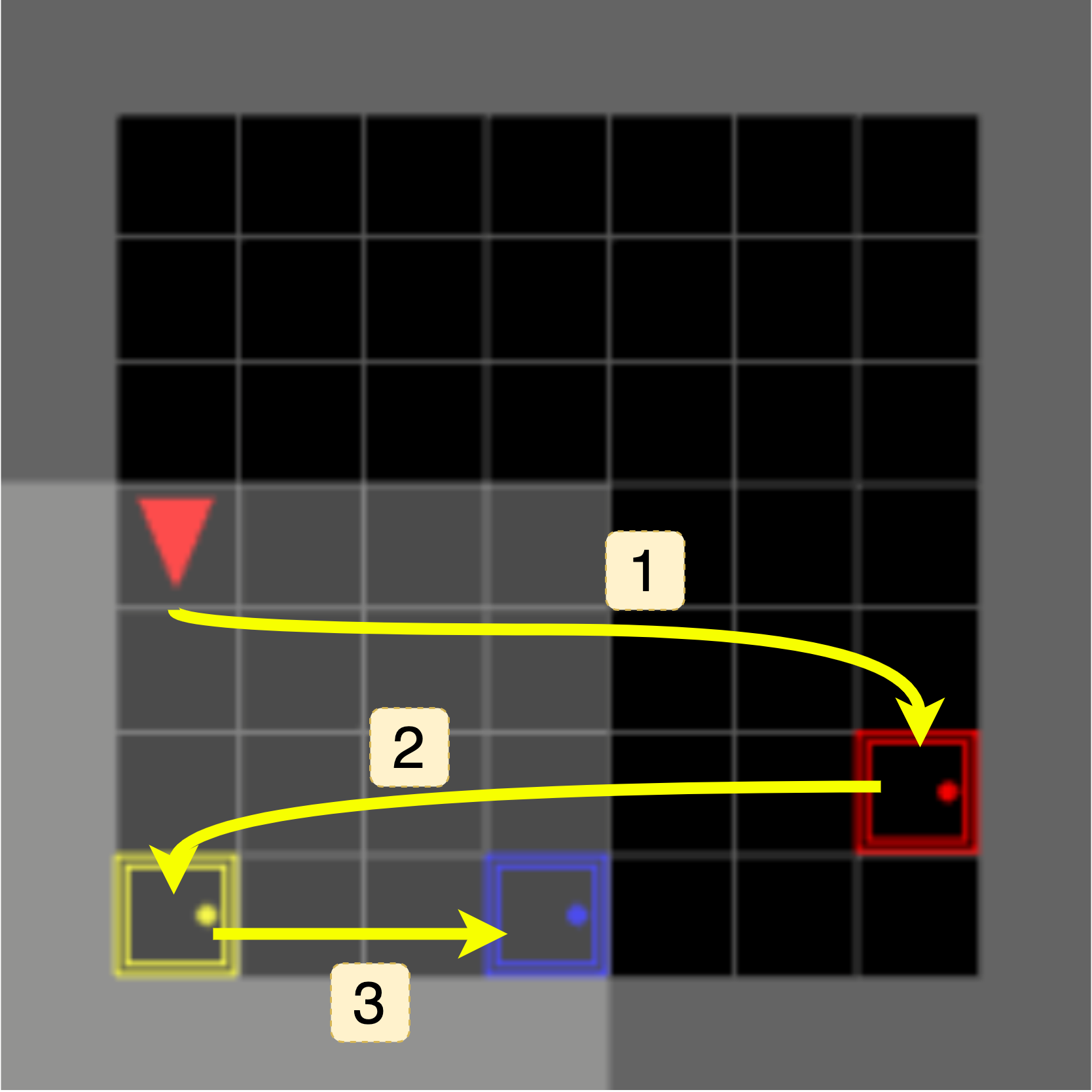}
    \caption{Illustration of the task and environment. The agent is required to navigate to and open three doors in a pre-defined sequence.}
    \label{fig:task}
\end{figure}

We evaluated our method in the Mini-Grid world environment \cite{chevalier2018minimalistic} on a multi-stage task. We recorded a few human task-solving interactions as a few-shot data set. 

The task is a simulation in a 2D grid world, as shown in Figure \ref{fig:task}. The final mission is to navigate the agent (the red arrow) to open three doors in different colors in a given sequence. The pre-trained sub-skills are opening one door of given color among doors of all other possible colors (all rgb combinations). The action space contains discrete choices, viz. $\mathcal{A}=$[move up, move down, move left, move right, pick, place, open]. If the agent opens the wrong door or takes a wrong action to the door, it would be punished with a -1 reward, and the task will reset. The initial position of the agent and all doors are randomly distributed. It receives a reward of 1 only after opening each door in the correct sequence. The maximum reward for each episode is 3 (maximum reward is 1 for sub-skills).  This environment is a simple and efficient abstraction of a multi-stage sparse reward task, in which sequence matters.

\paragraph{Results}
\begin{figure}[h]
    \centering
     \includegraphics[width=0.46\textwidth]{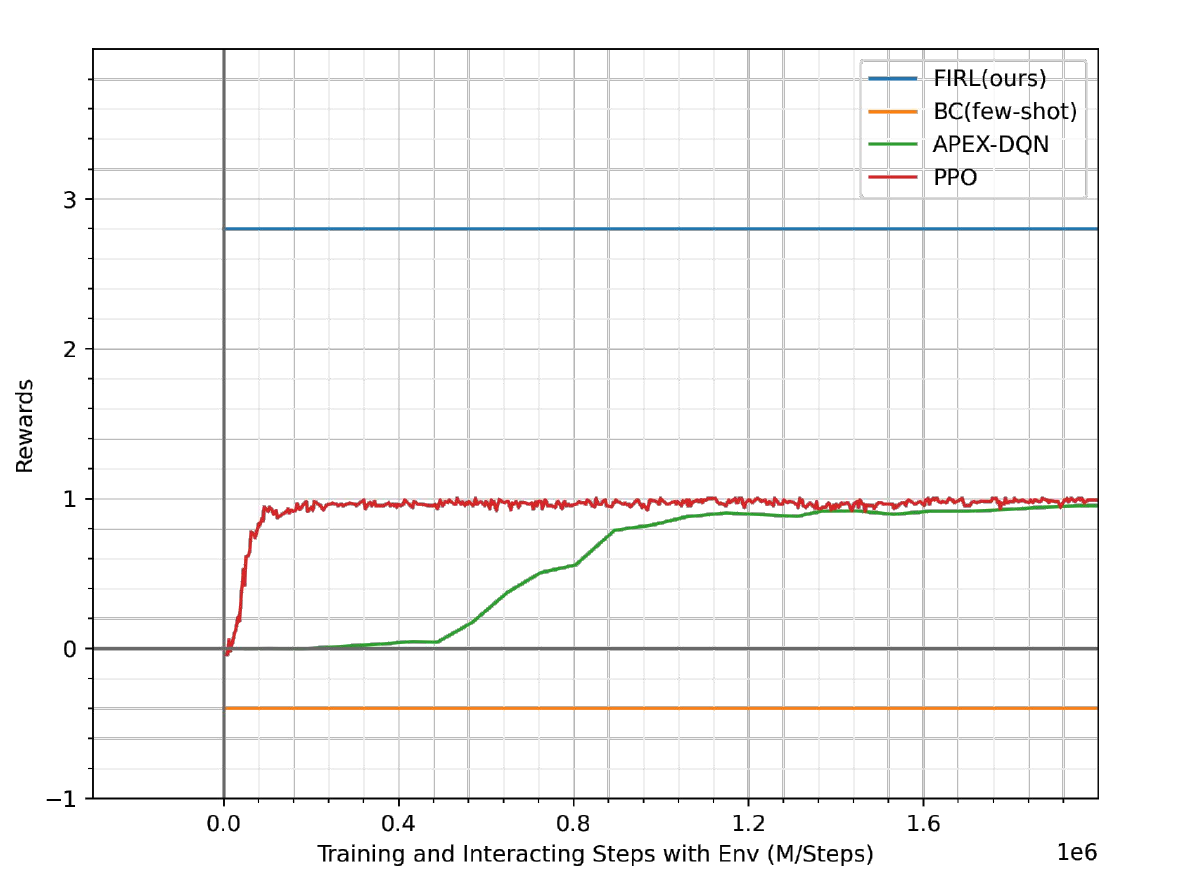}
    \caption{Learning curves of FIRL and baseline methods.}
    \label{fig:res}
\end{figure}

\begin{table*}
    \caption{Comparison of FIRL and related methods. \label{table:benchmark}}
    
    \begin{center}
    
    \resizebox{0.82\linewidth}{!}{
    
        \begin{tabular}{c||c|c|c|c}
            \hline
             & FIRL(ours) & PPO & APEX-DQN & BC \\
            \hline
            Average Rewards & $2.82\pm0.1$ & $0.96\pm0.03$ & $0.96\pm0.03$ & $-0.1\pm0.05$ \\
            \hline
            Success Rate (\%) & $94\pm3.0$ & 0 & 0 & 0 \\
            \hline
            Environment Train/Interaction Steps  & 0 & $\sim300,000$ & $\sim2,400,000$ & 0 \\
            \hline
            Give Number of Demos & 1 & 0 & 0 & 5 \\
            \hline
        \end{tabular}
    }
    \end{center}
\end{table*}
In the experiment, FIRL succeeded in the three-stage navigation and open-door task, with three corresponding sub-skills. It reached a success rate of $94\%$ after training on a single demonstration trajectory. The learning curves are shown in Figure \ref{fig:res}.
We compare learning performance on the task with the baseline algorithms PPO, APEX, and BC (few shots), and the results are listed in Table \ref{table:benchmark}. In particular, FIRL got the highest average reward of 2.82, while the others could only get 0.96 at maximum. Additionally, FIRL is the only method that completed the three-stage task, while PPO and APEX stopped exploring after the first reward. As for BC, it failed to master this few-shot task, as the information from five demos is far from enough. We did not include other baselines such as SPRIL \cite{pertsch2020accelerating} and RPL \cite{gupta2019relay}, as they require a much larger dataset, which is palpably not suitable for comparison with few-shot learning.

In summary, the FIRL algorithm outperformed the baselines in this multistage task. It exhibited the ability to instantly master a task from one-shot demonstrations with a high success rate. 

\paragraph{Ablation Study}
We experiment FIRL with or without mechanisms of Observation Minimization and Reward-augmented Loss. We define the following four experimental groups \textbf{FIRL(O)}, \textbf{FIRL(R)}, \textbf{FIRL(O+R)}, and \textbf{FIRL(\(\varnothing\))} (shown in Table \ref{table:ablation_groups}). 

\begin{figure}[h]
     \centering
     \begin{subfigure}{0.9\linewidth}
         \centering
         \includegraphics[width=\linewidth]{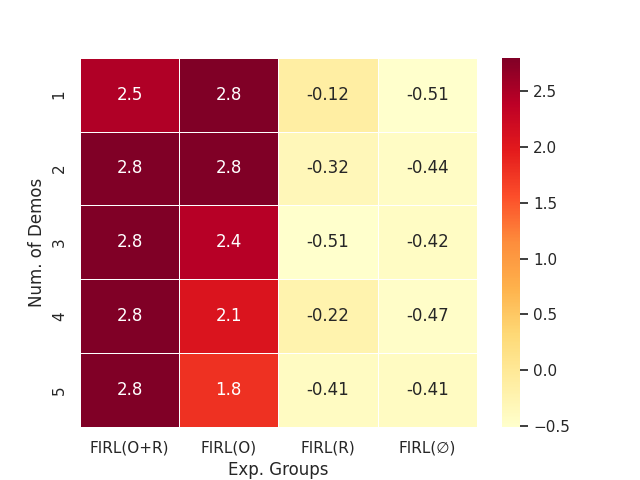}
         \caption{Average Rewards.}
         \label{fig:abla_rews}
     \end{subfigure}
     \hfill
     \begin{subfigure}{0.9\linewidth}
         \centering
         \includegraphics[width=\linewidth]{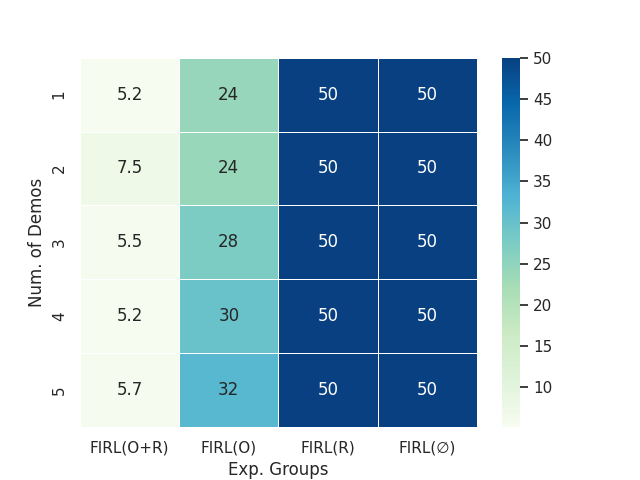}  
         \caption{Number of epochs required.}
         \label{fig:abla_epoch}
     \end{subfigure}
     \hfill
        \caption{Ablation study result. We examined four groups FIRL(O), FIRL(R), FIRL(O+R) and FIRL(\(\varnothing\)). A: Average reward for each group with different number of demonstrations. B: Number of training epochs required to achieve the same performance.}
\label{fig:ablation_al}
\end{figure}

\textbf{Observation Minimization:}
From Figure \ref{fig:abla_rews} and Figure \ref{fig:abla_epoch}, one can see the critical role of observation minimization (abbreviated by O) by comparing the FIRL (O + R) and FIRL (R) groups. It shows that observation minimization can significantly promote the average reward while reducing training time to a great extent. In summary, observation minimization is the key to achieving a high success rate and efficient sampling.
\begin{table}[H]
    \caption{Ablation Study groups. R: Reward-augmented loss. O: Observation minimization.}
    
    \label{table:ablation_groups}
    
    \begin{center}
    
    \resizebox{\linewidth}{!}{
    
    \begin{tabular}{c||c|c|c|c}
    \hline
     & FIRL(\(\varnothing\)) & FIRL(R) & FIRL(O) & FIRL(O+R) \\
    \hline
    Observation Minimization & \ding{55} & $\checkmark$ & \ding{55} & $\checkmark$ \\
    \hline
    Reward-augmented Loss& \ding{55} & \ding{55} & $\checkmark$ & $\checkmark$ \\
    \hline
    \end{tabular}
    }
    \end{center}
\end{table}
\textbf{Reward-augmented Loss:}
Albeit a crucial mechanism, it is the synergy of observation minimization and reward-augmented loss that achieves fast and reliable learning. In Figure \ref{fig:ablation_al}, it can be seen that although reward-augmented loss cannot greatly improve performance alone, it manages to largely increase efficiency after observation minimization is applied. 
In general, we can conclude that both mechanisms are necessary for FIRL to achieve its objective. Arguably, in the absence of the two mechanisms, the agent fails in a few shots of learning. 

\begin{figure}[b]
    \centering
     \includegraphics[width=0.46\textwidth]{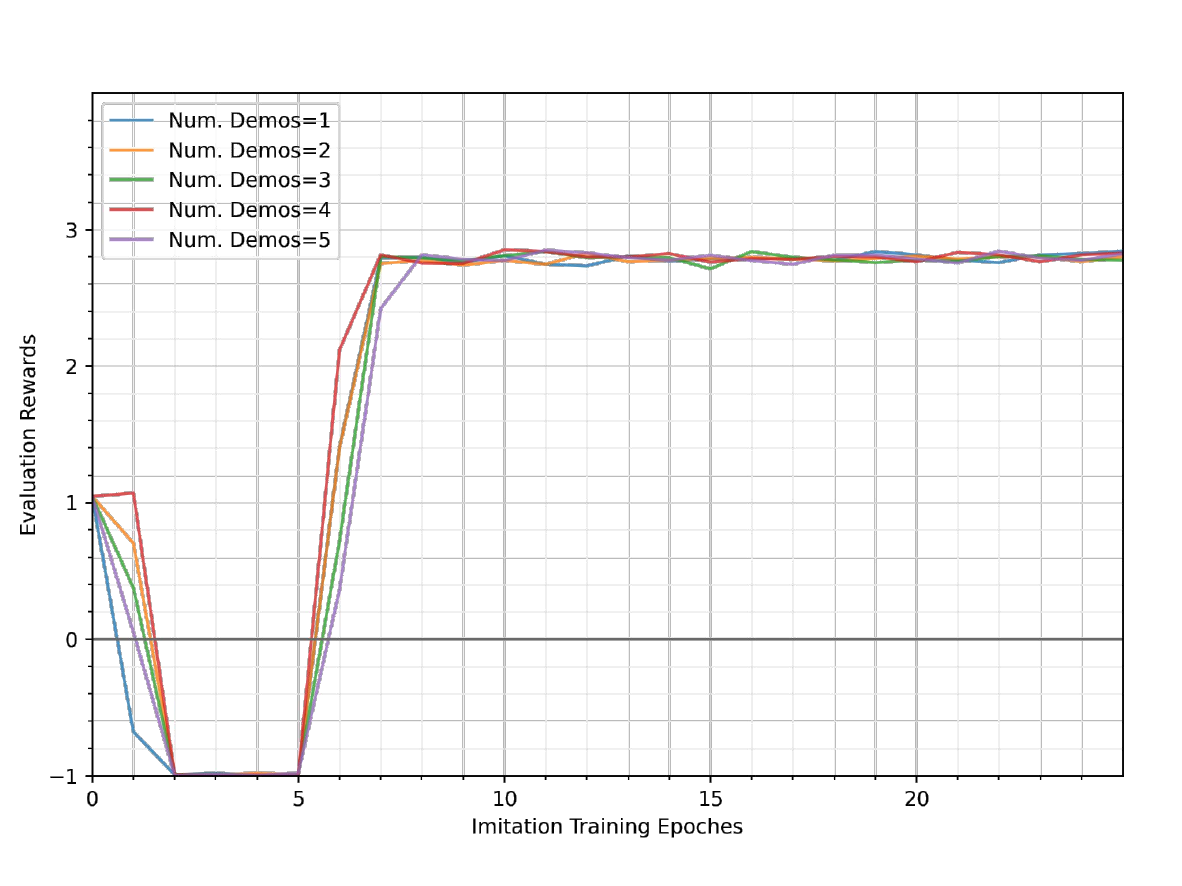}
    \caption{Learning curves of FIRL with different number of demonstrations.}
    \label{fig:demos}
\end{figure}

\textbf{Number of Demonstrations:}
We also examined the optimal number of demos required by FIRL (Figure \ref{fig:ablation_al}, Figure \ref{fig:demos}). By providing experiment groups with few shot demos in the range of 5, we found that the difference is nominal. It proves that FIRL can learn efficiently even with a single demonstration, meanwhile an increase number of demonstrations within few shot range is not necessary.

\section{Discussion}
In this work, we propose FIRL, an flexible algorithm for few-shot learning from human instructors. We introduced a hierarchical policy architecture for efficient policy reuse. With the rewards-augmented imitation
learning and observation minimization mechanisms, our agent can achieve extremely fast learning ability with a high success rate. The FIRL demonstrates a possible way of solving tasks possessing the features of: 1) Sparse reward, multi-stage and 
long horizon. 2) Flexible combination of skills. 3) Very few learning samples. 4) Requirement for lifelong learning. Thus, the FIRL can be viewed as a versatile agent, which also provides an idea for the long-term goal of building a general intelligent agent. It shows the potential to be an alternative solution for scenarios like Interactive Task Learning. 

Nevertheless, the FIRL is a first attempt under this architecture. There are still many aspects that could be improved. For example, we now select necessary features in observation minimization manually, while we hope to design an automated mechanism, or several standard operations for canonical situations. In addition, one may notice that we did not include methods of translating primitive human demonstrations into the trajectories that artificial agents can comprehend. As this is beyond the scope of this work, we hope to strengthen this part in future work. Lastly, we intend to further examine the performance of the proposed algorithm in more complex environments, e.g., 3D simulation or real robots, to verify
the validity of our design.

\bibliographystyle{ieeetr}
\bibliography{reference}
\end{document}